\documentclass[10pt,twocolumn,letterpaper]{article}

\usepackage[pagenumbers]{cvpr} 
\usepackage{times}
\usepackage{epsfig}
\usepackage{graphicx}
\usepackage{amsmath,amssymb}
\usepackage{multirow}
\usepackage{multicol}
\usepackage{arydshln}
\usepackage{booktabs}
\usepackage[accsupp]{axessibility} 
\usepackage{stmaryrd}
\usepackage{arydshln}
\usepackage{algorithm}
\usepackage{url}            
\usepackage{booktabs}       
\usepackage{amsfonts}       
\usepackage{nicefrac}       
\usepackage{microtype}      
\usepackage{xcolor}         
\usepackage{subfiles} 
\usepackage{adjustbox}
\usepackage{mathtools}
\usepackage{makecell}
\usepackage{multirow}
\usepackage[hypcap=false]{caption}
\usepackage{svg}            
\usepackage[noend]{algpseudocode}
\usepackage{listings}
\usepackage{pifont}
\usepackage[misc,geometry]{ifsym}

\newcommand{\sketch}{\mathcal{S}}

\newcommand{\renderer}{\mathcal{R}}

\newcommand{\z}[0]{\mathbf{z}}
\newcommand{\ours}{DreamWire}

\definecolor{xred}{RGB}{192,87,70}
\definecolor{xblue}{RGB}{42,183,202}
\definecolor{xgreen}{RGB}{106, 153, 78}

\newcommand{\keypoint}[1]{\vspace{0.12cm}\noindent\textbf{#1}\quad}
\newcommand{\cut}[1]{}

\usepackage{array}
\newcommand{\calcfactor}[1]{%
  \dimexpr#1\textwidth-2\tabcolsep-1.5\arrayrulewidth\relax
}
\newcolumntype{P}[1]{p{\calcfactor{#1}}}

\definecolor{cvprblue}{rgb}{0.21,0.49,0.74}
\usepackage[pagebackref,breaklinks,colorlinks,citecolor=cvprblue]{hyperref}

\usepackage[capitalize]{cleveref}
\crefname{section}{Sec.}{Secs.}
\Crefname{section}{Section}{Sections}
\Crefname{table}{Table}{Tables}
\crefname{table}{Tab.}{Tabs.}

\crefformat{footnote}{#2\footnotemark[#1]#3}

\usepackage{xr}

\def\paperID{7} 
\def\confName{CVPR}
\def\confYear{2024}

\begin{document}

\title{Wired Perspectives: Multi-View Wire Art Embraces Generative AI\vspace{-0.5cm}}

\author{%
  Zhiyu Qu$^{1}$\:\; Lan Yang$^{2}$\:\; Honggang Zhang$^{2}$\:\; Tao Xiang$^{1}$\:\; Kaiyue Pang$^{1}$\:\; Yi-Zhe Song$^{1}$\\[0.2cm]
  $^{1}$SketchX, CVSSP, University of Surrey \quad
  $^{2}$Beijing University of Posts and Telecommunications \\
  \small{\texttt{\{z.qu, t.xiang, k.pang, y.song\}@surrey.ac.uk    }}    
  \small{\texttt{\{ylan, zhhg\}@bupt.edu.cn}} \vspace{0.1cm}\\
  \url{https://dreamwireart.github.io}
}

\twocolumn[{
    \renewcommand\twocolumn[1][]{#1}
    \maketitle
    \begin{center}
        \centering
        \vspace{-0.7cm}
        \includegraphics[width=0.99\textwidth]{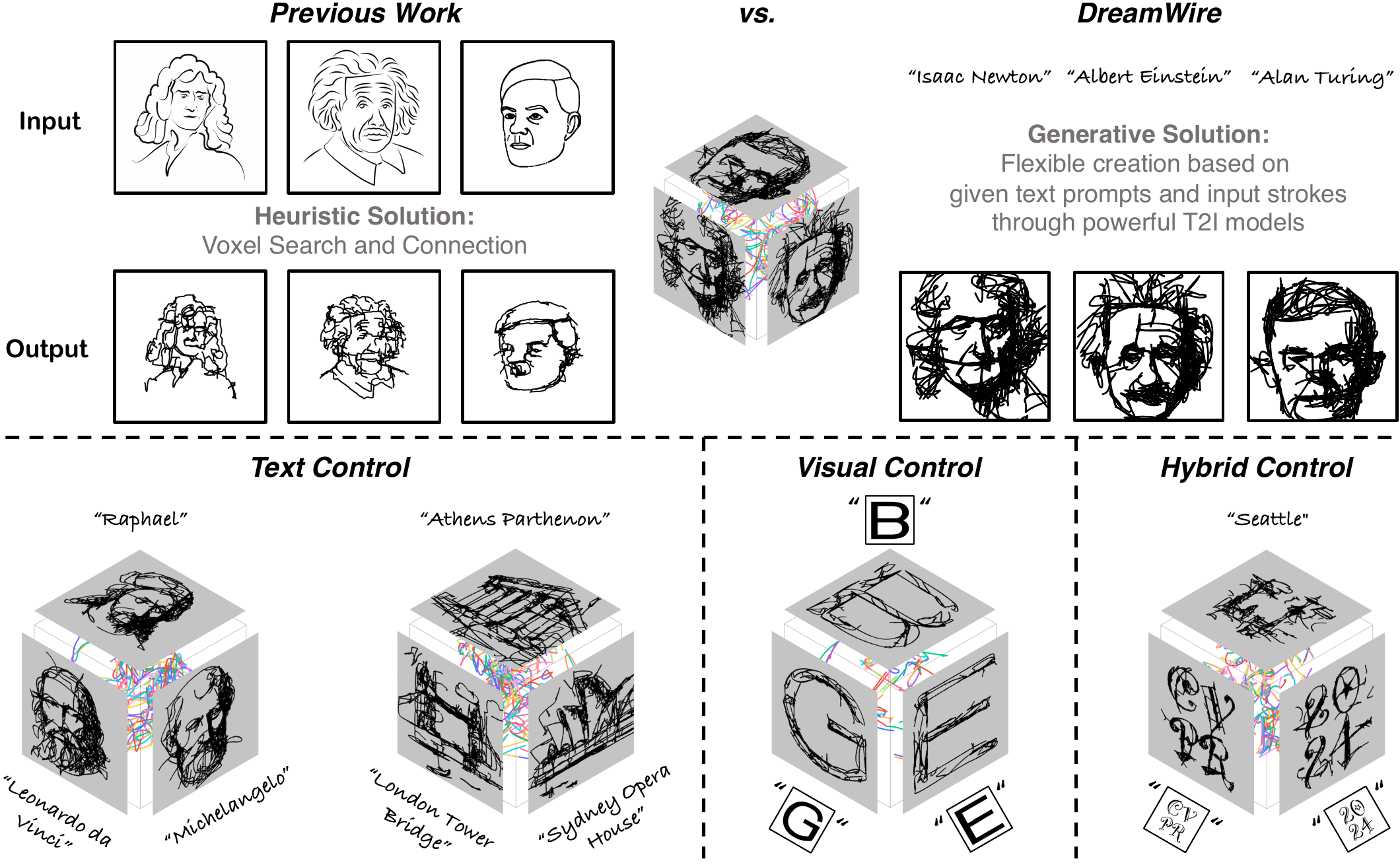}
        \vspace{-0.1cm}
        \captionof{figure}{\textbf{Multi-view art done generatively.} We present \textit{{\ours}} as the first system that takes user prompt for each view as input -- either via the expressive vehicle of text or image -- and produces 3D line sculptures showing distinct interpretations when viewed at different angles, \ie, multi-view wire art (MVWA). Compared to previous rule-based work, we significantly improve the quality of MVWA by utilising the flexible drawing capabilities of a universal generative prior (diffusion models or ControlNet). Notably, the ``GBE'' here pays tribute to the book ``Gödel, Escher, Bach: an Eternal Golden Braid'' \cite{GEB}, which discusses how systems can acquire meaningful context despite being made of ``meaningless'' elements, just like what MVWA does.}
        \label{fig:overview}
        \vspace{-0.2cm}
    \end{center}
}]


\begin{abstract}
\vspace{-0.4cm}
Creating multi-view wire art (MVWA), a static 3D sculpture with diverse interpretations from different viewpoints, is a complex task even for skilled artists. In response, we present \ours, an AI system enabling everyone to craft MVWA easily. Users express their vision through text prompts or scribbles, freeing them from intricate 3D wire organisation. Our approach synergises 3D B\'ezier curves, Prim's algorithm, and knowledge distillation from diffusion models or their variants (\eg, ControlNet). This blend enables the system to represent 3D wire art, ensuring spatial continuity and overcoming data scarcity. Extensive evaluation and analysis are conducted to shed insight on the inner workings of the proposed system, including the trade-off between connectivity and visual aesthetics. 
\end{abstract}
\vspace{-0.7cm}
\section{Introduction} \label{sec:intro}
\vspace{-0.2cm}
\textit{A great thought begins by seeing something differently, with a shift of the mind’s eye.}

\par\noindent\rule{\linewidth}{0.4pt}

\hfill{\textit{Albert Einstein}}

There is an artist in everyone, they say. Attending an art exhibition, being mesmerised by a 3D wire-art installation from Matthieu Robert-Ortis\footnote{\url{https://leonacreo.com/sculptures-by-matthieu-robert-ortis/}} is what motivated this paper! As an AI practitioner, the immediate question was, ``Can I program this?'' Not to replace artists, but rather, for fun, for finding that artist within myself, and for the vision of democratising art creation for everyone!

Lighthearted as it might sound, this endeavour holds scientific value on two fronts. First, it delves into the uncharted territory of wired-art generation using current generative AI \cite{rombach2022high, midjourney, shi2020improving}, seeking to understand the limits of these technologies in the realm of this unique artistic form. Secondly, it contributes to the ongoing dialogue by exploring the expansion of existing 2D-focused generation methods into the intricate domains of 3D and perhaps more challengingly, the extreme abstraction presented by wire art.

Multi-view wire art (MVWA) \cite{hsiao2018multi} is a unique form of art that leverages wire as a flexible medium to create complex 3D objects, whereupon different viewpoints, multiple interpretable images appear -- recall those 2D pictures where you move your head and see different things. This time, you are walking around a 3D installation, and upon different viewing angles, you see different 2D depictions (see Fig.~\ref{fig:overview}). Being prohibitively difficult for novice users, creating MVWA is an extremely time-consuming task even for qualified artists. Apart from artistic ideation, working with reverse projection (2D to 3D), efforts have been made on physics so the installation does not collapse. Our ambition for MVWA, one that focuses on democratising its creation for everyone, is removing all said challenges but limiting its creation to just ideation (perhaps not entirely artistic, though!). That is, specifying what you want each view to look like, and bingo -- the final 3D art form!

We present a system named \textit{{\ours}} to do just that. All users need to generate 3D wire art is a set of text prompts (\eg, ``a portrait of Einstein'') or rough scribbles (\eg, styled writings of ``CVPR''), each for a 2D view. Fig.~\ref{fig:overview} illustrates some examples, and for a more immersive experience, we offer fully interactive MVWA demos in our project page -- please do set your eyes on them; we promise they won't be boring! However, there is a caveat: there is an upper bound (\textit{three}) on how many viewpoints an MVWA object could support, largely due to the degree of conflict in the 3D wire space that a large number of views would introduce.

\begin{figure*} 
    \centering
    \includegraphics[width=\linewidth]{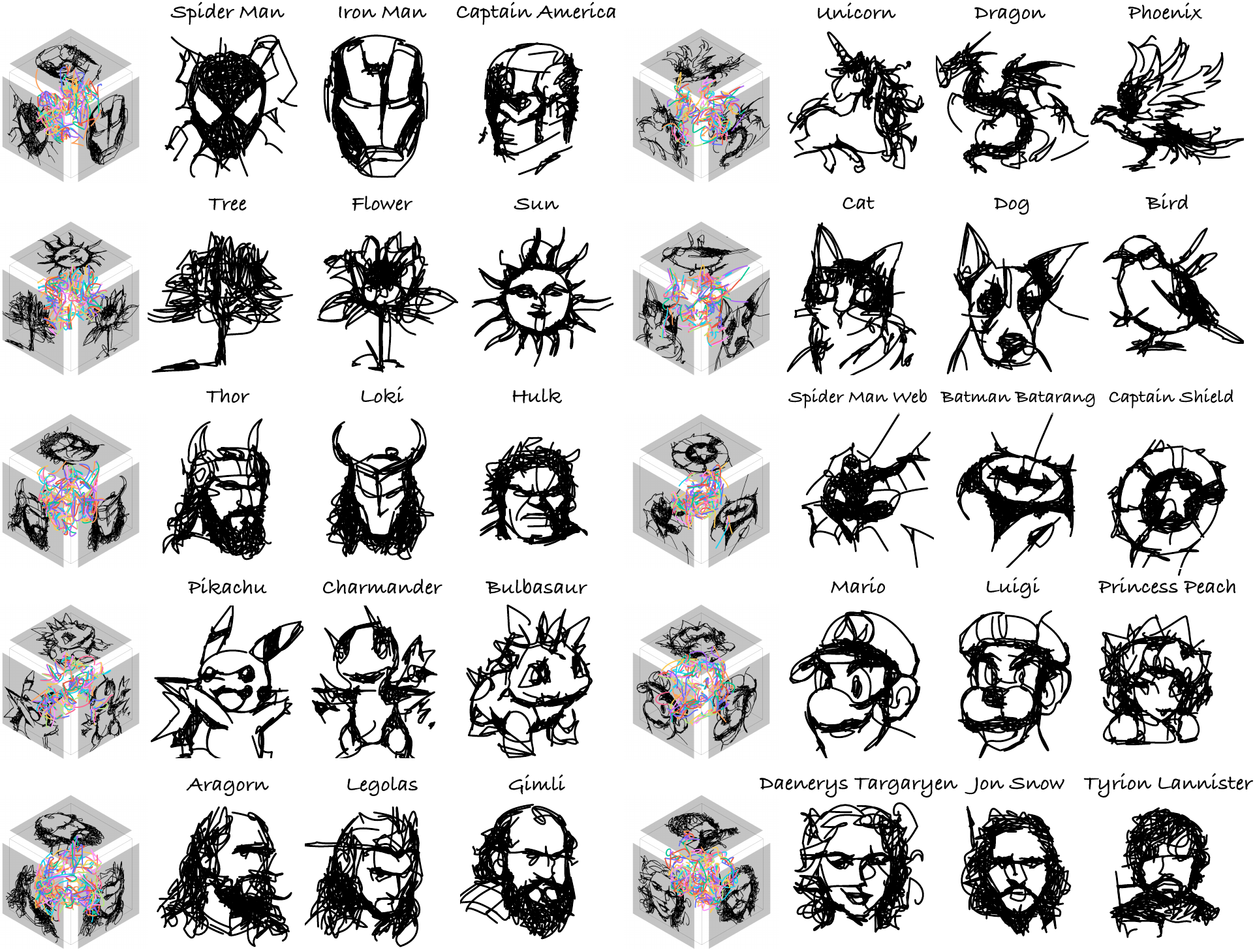}
    \caption{\textbf{MVWA generated via {\ours}}. The textual prompts employed predominantly include ``a head of [character]'' and ``a simple drawing of [item]''. Notably, all captions for these MVWA have been sourced from ChatGPT \cite{chatgpt}. We prompt it to return three major subjects of interests under a given topic, \eg, three celebrated movie characters of the United States of America.} 
    \label{fig:showtime}
    \vspace{-0.2cm}
\end{figure*}

Computational methods for MVWA \cite{hsiao2018multi} or related art forms alike \cite{shadowart} have been attempted before but only appear as rule-based endeavours -- they rely on a set of prewritten rules to construct an MVWA piece\footnote{For readers unfamiliar with the existing ``assembly manual'' for MVWA, we briefly summarise the rules here: i) back-project the 2D images to 3D via generalised cones and discrete the intersection of the camera’s viewing frustums with a fixed resolution of vocalisation; ii) inevitably, some of these initial voxels only represent the line image from their own source, resulting in inconsistent visual impacts on other viewpoints. To address this, optimisation of a voxel displacement problem is needed, whereby conflicting voxels are either merged into one or smoothed with neighbouring voxels as a more holistic visual entity. iii) Voxels are subjected to further manipulations, often targeting issues more delicate than inconsistency, including redundancy, complexity, quality, etc.}. Much like any rule-based methods for vision problems (SIFT, HOG), these approaches are advantageous for their full transparency of the playbook but fall short in generalisation. This is discussed in Sec.~\ref{subsec:main_results}, where existing approaches, guided by human-informed rules, can create MVWA pieces whose 2D projections align perfectly with user inputs, but they collapse when faced with slightly more complex combinations of 2D view images. Reproducing most of the results shown in Fig.~\ref{fig:showtime} would therefore be a stretch because there is not yet a rule-based system that can generate arbitrary plausible 2D visual images from a text string, let alone generating MVWA on top of that.

We face two key challenges: (i) how to represent 3D wire art while ensuring connectivity (so it does not collapse!), and (ii) how to ensure effective learning from extremely scarce MVWA training examples. For the former, we leverage 3D B\'ezier curves to solve a connectivity (``wiredness'') problem that cannot be easily achieved in a naive way, such as by chaining control points (see Fig.~\ref{fig:oneline}). Instead, we treat each B\'ezier curve independently and propose a loss function to spatially constrain their degree of freedom. At each iteration, we depict the currently learned wires as a weighted undirected graph and apply Prim's algorithm \cite{prim1957shortest} to derive a subset of edges (including all vertices) corresponding to a minimum spanning tree. The spatial continuity of wires is thus assured by minimising the distance between each parent and child vertex. On the latter challenge, we opt for \textit{per}-instance learning and base generalisation on knowledge distillation from a powerful generative visual prior (diffusion models \cite{rombach2022high} or their variant, \ie, ControlNet \cite{zhang2023adding} in this case). In unison, our system begins with a set of randomly initialised B\'ezier curves, which, after 2D projection and vector-to-raster conversion, are fed into diffusion models to match the user target text or image and updated via the typical score distillation sampling (SDS) \cite{poole2022dreamfusion} process. 

In summary, our contributions are threefold: (i) empowering everyone to become a wired 3D (MVWA) artist (even if only half-decent), and scientifically, (ii) employing B\'ezier curves and Prim's algorithm to represent 3D wire art, and (iii) utilising a powerful generative visual prior through a designed rendering strategy to overcome data scarcity and the limitations of rule-based methods.

\section{Related work} \label{sec:related_work}

\keypoint{Vector Graphics.} Scalable vector graphics (SVGs), in contrast to images composed of raster pixels, are defined by extensible markup language (XML) covering lines, shapes, or curves. B\'ezier curve is one of the most pronounced SVG formats, which relies on a set of ``control points'' to define a smooth line segment. While there is no denying that computer vision has predominantly invested in understanding raster images, recent efforts witnessed several important breakthroughs in the generative modelling of SVGs, mostly for B\'ezier curves. B\'ezierSketch \cite{das2020beziersketch} first introduced an inverse graphics approach to sketch stroke embedding that trains an encoder to embed each stroke to its best fit B\'ezier curve, and their subsequent work extends this idea to a more generalised case with variable-degree B\'ezier control. Another line of works \cite{Li:2020:DVG, CLIPDraw, styleclipdraw, vinker2022clipasso} directly utilise B\'ezier curves to govern the general-purpose vector graphic generation process. VectorFusion \cite{jain2022vectorfusion} employs diffusion models as transferable priors to generate high-quality abstract vector graphics from text captions. SketchDreamer \cite{qu2023sketchdreamer} presents an interactive method for text-driven vector sketch generation, adeptly incrementing strokes to an initial vector sketch in accordance with a user-specified text prompt. These works however only contribute to the application of 2D B\'ezier curves. We consider the problem of how to render 3D B\'ezier curves using a 2D B\'ezier renderer \cite{Li:2020:DVG}, which is significantly different from previous work.

\keypoint{Diffusion Prior.} There has been a burgeoning interest in denoising diffusion probabilistic models \cite{ddpm, ddim, du2023demofusion, ng2023dreamcreature}, also known as score-based generative models \cite{song2019gradients, song2020score}, thanks to the remarkable generative prowess they have shown. Consequently, an increasing number of studies \cite{ddpmpnp, poole2022dreamfusion, singer2023make, lin2023magic3d, tang2023make, han2023headsculpt} emerge as to how to leverage pretrained diffusion models to act as effective visual priors for generative supervision. DDPMPnP \cite{ddpmpnp} introduces a partitioning of diffusion models into a base prior and a conditional constraint, enabling versatile applications in perceptual tasks like conditional image generation and segmentation. DreamFusion \cite{poole2022dreamfusion} optimises NeRF parameters using an efficient, high-fidelity Score Distillation Sampling (SDS) loss, facilitated by a 2D diffusion image prior to text-to-3D synthesis. Make-A-Video \cite{singer2023make} employs spatial-temporal modules built on 2D text-to-image diffusion models, realising text-to-video generation without the need for paired samples. We adopt the idea of applying the diffusion prior to a differentiable image parameterisation \cite{mordvintsev2018differentiable} (DIP) as proposed by DreamFusion, with the difference that we focus on the generation of multi-view wire art.

\keypoint{Multi-View Art.}
Multi-View Art entails the presentation of multiple perspectives or views within a singular artwork \cite{oliva2006hybrid, kuo2016generating, bermano2012shadowpix, baran2012manufacturing, sela2007generation,hsiao2018multi}. The techniques for achieving varied visual perceptions in an artwork can span a range of approaches: from altering the viewing distance \cite{oliva2006hybrid}, adjusting the viewing direction \cite{hsiao2018multi, sela2007generation}, to changing illumination from specific directions \cite{shadowart, alexa2010reliefs, bermano2012shadowpix}. The underlying factors prompting such phenomena are multifaceted, including the use of optical materials \cite{zeng2021lenticular, perroni2023constructing}, innovative structural design \cite{hsiao2018multi, shadowart, bermano2012shadowpix}, or specialised devices \cite{hosseini2020portal}. We unprecedentedly introduce powerful text-to-image generation models to this problem, elevating the upper limit of creativity and simplifying and democratising the art creation process.

\section{Methodology} \label{sec:formulation}

\subsection{Differentiable 3D MVWA rendering} \label{subsec:render} 

We represent a 3D multi-view wire art $\sketch$ as a set of individual wires $\{s_1, \cdots, s_n\}$. Each individual wire employs a cubic 3D B\'ezier curve, which is rigorously defined by a quartet of 3D control points $\{p_0, p_1, p_2, p_3\}$, detailed in Eq.~\ref{eq:bezier}:
\begin{equation} \label{eq:bezier}
\begin{aligned}
    B(t) = (1 - t)^3p_0 + 3(1 - t)^2t p_1 + 3(1 - t)t^2 p_2 + t^3 p_3 ,
\end{aligned}
\end{equation} 
\noindent where $t\in[0,1]$. A straightforward approach to render 3D B\'ezier curves is to consider a specific plane and calculate the projection of every point along the curve onto this plane. Given a plane $\pi$ characterised by its normal vector $N$, the projection of the 3D cubic B\'ezier curve $B(t)$ onto $\pi$ is formulated in Eq.~\ref{eq:bezier'}:
\begin{equation} \label{eq:bezier'}
\begin{aligned}
    &B^{\prime}(t) = B(t) - [N \cdot (B(t) - q)] N ,
\end{aligned}
\end{equation}
where $q$ is an arbitrary point on $\pi$. However, prevalent 2D B\'ezier rendering techniques, such as the one discussed by \cite{Li:2020:DVG}, as well as 3D point cloud rendering tools \cite{liu2019softras, ravi2020pytorch3d, Laine2020diffrast}, do not support the rendering of such 3D B\'ezier curves in Eq.~\ref{eq:bezier'}. We propose an inquiry: Is it feasible to render 3D B\'ezier curves utilising existing 2D B\'ezier curves renderers? The answer is \textit{YES}.

Our objective is to prove that \textit{the projection of a 3D B\'ezier curve onto a plane}, denoted as $B^{\prime}(t)$, is equivalent to \textit{the 2D B\'ezier curve, whose control points are the projections of the original control points} of $B(t)$ onto the same plane, expressed as $B^{\prime\prime}(t)$. Utilising Eq.~\ref{eq:bezier'}, the 2D B\'ezier curve $B^{\prime\prime}(t)$ formed by the projection points of $p_i$ on $\pi$ can be expressed as, 
\begin{equation} \label{eq:bezier''}
    B^{\prime\prime}(t) = (1 - t)^3 p^{\prime}_0 + 3(1 - t)^2t p^{\prime}_1 + 3(1 - t)t^2 p^{\prime}_2 + t^3p^{\prime}_3.
\end{equation}

The equivalence of $B^{\prime}(t)$ and $B^{\prime\prime}(t)$ can be systematically demonstrated by applying the principles outlined and the property of vector addition to expand and transform Eq.~\ref{eq:bezier''}:
\begin{equation} \label{eq:equal}
    \begin{aligned}
      B^{\prime\prime}(t) 
      &=  \underbrace{(1 - t)^3p_0 + 3(1 - t)^2t p_1 + 3(1 - t)t^2 p_2 + t^3 p_3}_{B(t)} -\\
      & \{N \cdot [\underbrace{(1 - t)^3 p_0 + 3(1 - t)^2t p_1 + 3(1 - t)t^2  p_2 + t^3 p_3}_{B(t)} \\
      & - q]\}N = B(t) - [N \cdot (B(t) - q)] N = B^{\prime}(t).
    \end{aligned}
\end{equation}
\noindent This proof enables us to reframe the challenge of rendering 3D B\'ezier curves as essentially a 2D rendering task, anchored in the projection of 3D control points. Consequently, we are able to directly optimise the 3D wire art $\sketch$ using a differentiable 2D B\'ezier curve renderer, \ie, DiffVG \cite{Li:2020:DVG}.

\begin{figure*} 
    \centering
    \includegraphics[width=0.96\linewidth]{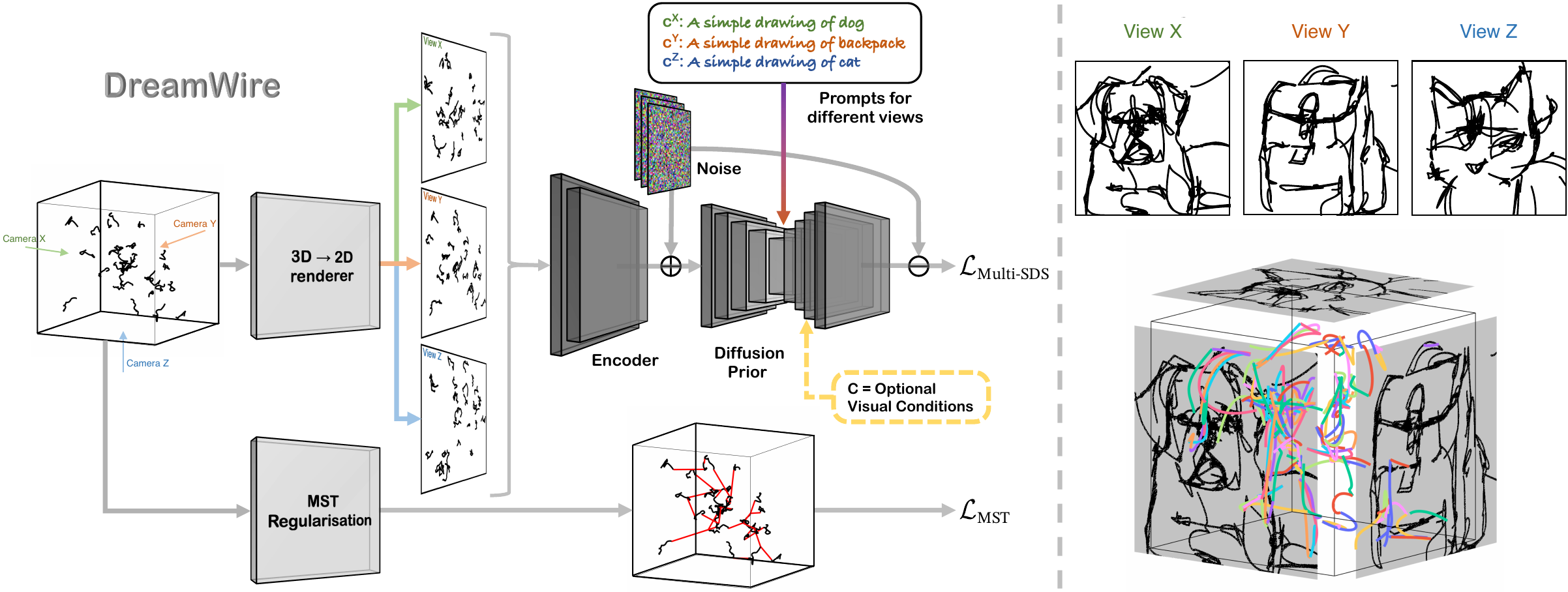}
    \vspace{-0.1cm}
    \caption{\textbf{Schematic overview of {\ours}.} Starting from an initial set of random 3D B\'ezier curves, we project these curves onto a given 2D plane and process them into normal raster images. It follows that these images are fed into a generative diffusion model and optimised towards a visual target. In addition, we use the MST algorithm to constrain the distance between curves. Here we present a MVWA sample output under the condition $\{c^X, c^Y, c^Z\}=\{$``dog'', ``backpack'', ``cat''$\}$.  
    }
    \label{fig:framework}
    \vspace{-0.2cm}
\end{figure*}

\subsection{{\textbf{\ours}}} \label{sec:pipeline}
The overall pipeline of our \textit{\ours} is depicted in Fig.~\ref{fig:framework}. Users just need to provide three distinct inputs $c=\{c^X, c^Y, c^Z\}$, corresponding to projections from three mutually orthogonal viewpoints $\{X, Y, Z\}$. The primary objective of our MVWA generation system is to produce a 3D wire art, $\sketch$, such that its projections onto each of these viewpoints align with the user’s specified inputs. 

Initially, we initialise a 3D wire art $\sketch = \{s_i\}_{i=1}^n$, where the control points of each wire are randomly initialised. We define the three planes of projection as $\pi^X, \pi^Y, \pi^Z$, with their corresponding normal vectors $N^X, N^Y, N^Z$, which relate to the three user-provided viewpoints. Utilising Eqs.~\ref{eq:bezier''} and \ref{eq:equal}, we can determine the projection of $\sketch$ on plane $\pi^X$ as,
\vspace{-0.4cm}
\begin{equation}
\begin{aligned}
    \sketch^X &= \{\hat s_i^X\}_{i=1}^n = \{\hat B_i^X(t)\}_{i=1}^n , \\
    \text{where} \quad \hat B_i^X(t) &= \sum\nolimits_{j=0}^{3} \binom{3}{j} (1-t)^{3-j} t^j \hat p^{Xj}_i , \\
    \hat p^{Xj}_i &= p^j_i - [N^X\cdot(p^j_i - q^X)]N^X ,
\end{aligned}
\end{equation}
\noindent where $q^X$ is any point on plane $\pi^X$. We then utilise a differentiable 2D B\'ezier curve renderer, denoted as $\renderer$, to produce the rasterised projection. These projections are subsequently processed through the encoder $E_\phi$ of a Latent Diffusion Model (LDM) \cite{rombach2022high}, utilising the Score Distillation Sampling (SDS) loss \cite{poole2022dreamfusion} to estimate $\mathbf{z}^X = E_\phi(\renderer(\sketch^X))$. During each forward diffusion timestep, we introduce random noise to the latents, $\mathbf{z}^X_t = \alpha_t \mathbf{z}^X + \sigma_t \boldsymbol{\epsilon}$, and apply the teacher model $\hat{\boldsymbol{\epsilon}}_\phi(\mathbf{z}^X_t; t)$, conditioned on $c^X$, for denoising. This process is replicated for $\{Y, Z\}$. The optimisation targets all control points (collectively represented as $\mathbf{P}$), and is steered by the SDS loss, as expressed in Eq.~\ref{eq:multi-sds}:
\begin{equation}\label{eq:multi-sds}
\resizebox{0.48\textwidth}{!}{
$\begin{aligned}
    \nabla_\mathbf{P} \mathcal{L}_\text{Multi-SDS} &=
    \mathbb{E}_{t, \epsilon} \left[ w(t) \Big(\hat{\epsilon}_\phi(\alpha_t \z_t^X + \sigma_t \epsilon; c^X, t)  - \epsilon\Big) \frac{\partial \z^X}{\partial \mathbf{P}} \right] \\
    &+\mathbb{E}_{t, \epsilon} \left[ w(t) \Big(\hat{\epsilon}_\phi(\alpha_t \z_t^Y + \sigma_t \epsilon; c^Y, t)  - \epsilon\Big) \frac{\partial \z^Y}{\partial \mathbf{P}} \right] \\
    &+\mathbb{E}_{t, \epsilon} \left[ w(t) \Big(\hat{\epsilon}_\phi(\alpha_t \z_t^Z + \sigma_t \epsilon; c^Z, t)  - \epsilon\Big) \frac{\partial \z^Z}{\partial \mathbf{P}} \right] .
\end{aligned}$}
\end{equation}

Here, $w(t)$ represents the weighting function, and $t\in [1, 2, \cdots, T]$ denotes the timestep. To afford users greater flexibility in input types, we adopt the multi-conditional diffusion model approach, as suggested by \cite{qu2023sketchdreamer}, utilising a ControlNet~\cite{zhang2023adding} to govern the diversity and guide the diffusion model generation processes. This results in a controllable variant of Eq.~\ref{eq:csds}:
\begin{equation} \label{eq:csds}
    \nabla_\mathbf{P} \mathcal{L}_\text{CSDS} = \mathbb{E}_{t, \epsilon}\left[w(t)\left(\hat\epsilon_\phi(\alpha_t \z_t + \sigma_t \epsilon; c, t, C)  - \epsilon\right) {\frac{\partial \z}{\partial \mathbf{P}}}\right] ,
\end{equation}
\noindent where $C$ denotes visual conditions for ControlNet, \eg, canny edges, HED boundaries, user scribbles, human poses, semantic maps, depths, \textit{etc}. With the capabilities of ControlNet, users' input conditions can extend beyond text captions of visual concepts to include spatial layouts, enabling personalised customisation.

\subsection{MVWA to Reality}

Technical approach laid out in Sec.~\ref{sec:pipeline} only allows a digital construct of MVWA instance within the AR/VR environment. To enable a real tangible MVWA entity in real life that respects the law of physics however remains highly challenging. The wires ${s_1, s_2, \ldots, s_n}$ we acquire so far are incapable of sustaining stability in suspension; they mandate methodical interconnections to cultivate a stable and supportive structure. Taking inspiration from \cite{hsiao2018multi},  we approach this challenge by framing it as a classic minimum spanning tree (MST) problem, with the isolated wires and their spatial relationships represented as a graph. For a set of $n$ wires $\{s_1, s_2, \ldots, s_n\}$ and $\mathbf{P}$ representing all control points, we introduce $\ddot{\mathbf{P}}$ to denote the endpoints of all wires. For any pair of wires $\{s_i, s_j\}$, we calculate the Euclidean distance between their endpoints in four different ways\footnote{ $\mathcal{E}_{ij} = min(\parallel p^0_i - p^0_j\parallel^2, \parallel p^3_i - p^0_j\parallel^2, \parallel p^3_i - p^3_j\parallel^2, \parallel p^0_i - p^3_j\parallel^2)$}, electing the smallest of these as $\mathcal{E}_{ij}$. Through the assessment of the Euclidean distances amongst all endpoints, we proceed to construct a densely interconnected undirected graph $\mathcal{G}$, comprising $n$ vertices represented as $\{s_1, s_2, \ldots, s_n\}$, with the edges bearing weights equivalent to $\mathcal{E}_{ij}$.

\begin{figure} 
    \centering
    \includegraphics[width=\linewidth]{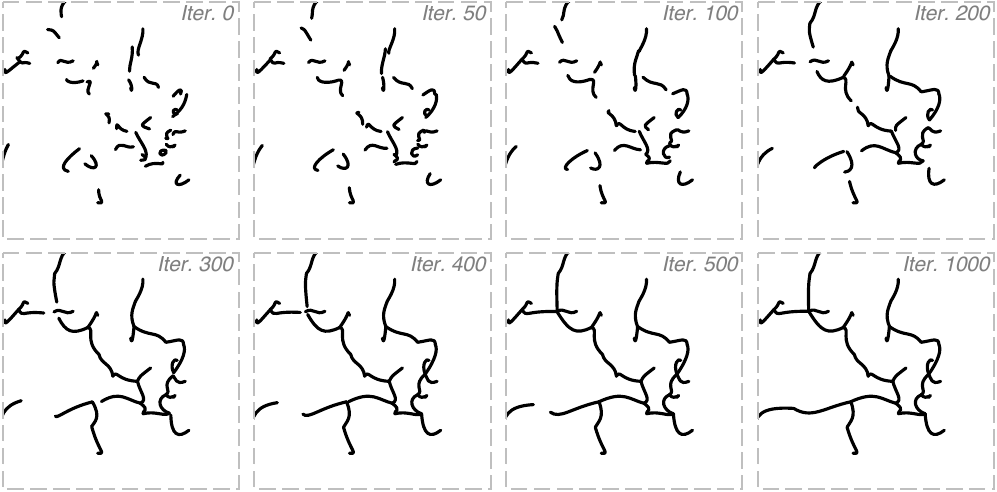}
    \caption{Effect of the MST regularisation on a set of randomly initialised B\'ezier curves.
    }
    \label{fig:tree}
    \vspace{-0.2cm}
\end{figure}

Our objective is to identify a subset of the edges of $\mathcal{G}$ that binds all the wires together, forming a cycle-free structure with the minimum aggregate edge weight. Employing Prim’s Algorithm \cite{prim1957shortest}, we derive the minimum spanning tree, and the associated cost is formulated as follows:
\begin{equation}
    \mathcal{L}_\text{MST}({\ddot{\mathbf{P}}}) = \sum Prim(\mathcal{E}_{ij}) \quad i,j \in [1, n] .
\end{equation}

Consequently, our final training objective is articulated as:
\begin{equation}
    \mathcal{L} = \mathcal{L}_\text{Multi-SDS}(\mathbf{P}, c) + \lambda * \mathcal{L}_\text{MST}({\ddot{\mathbf{P}}}).
\end{equation}
\noindent Here, $\lambda$ functions as a balancing factor between aesthetic appeal and structural realism. 
In Fig.~\ref{fig:tree}, we present a qualitative depiction of the impact elicited by the MST regularisation, effectively demonstrating how wires, initially scattered, progressively coalesce into a stable and integrated structure.

\section{Experiments} \label{sec:experiment}
\subsection{Settings}
\keypoint{Implementation.} Building upon the methodologies employed in \cite{jain2022vectorfusion} and \cite{qu2023sketchdreamer}, we initiate each 3D B\'ezier curve of MVWA comprising 5 segments, maintaining a constant width and adopting a uniform black colour. Prior to inputting the 2D projection of the MVWA into the diffusion model, we employ random affine augmentations (RandomPerspective and RandomResizedCrop) to refine the projection's quality and to reinforce the optimisation process against the potential adversarial examples. To optimise the MVWA, we employ the Adam optimiser~\cite{adam2014kingma} across 2000 iterations, setting the learning rate to 1. Within our configuration, we adopt a guidance scale equal to 100. All experiments are executed on an NVIDIA A100 GPU.

\keypoint{Data preparation.} 
For the visual control setting, to ensure a fair comparison, the input sets employed are identical to that of the baseline methods \cite{shadowart, hsiao2018multi}. For the text control setting, a selection of 96 daily item categories was randomly drawn from the QuickDraw dataset \cite{sketchrnn}. Each category name was then inserted into a standard template as ``a simple drawing of [item]'' and these were subsequently randomised into 32 distinct input sets.

\keypoint{Evaluation metrics.} 
Within the text control setting, we utilise the CLIP \cite{clip} score and R-Precision \cite{Park2021BenchmarkFC} as metrics to assess the similarity between the input text condition and 2D rasterised projection of the synthesised MVWA. For visual control, DINO \cite{Caron2021EmergingPI} is employed to quantify the similarity between the 2D rasterised projection of the generated MVWA and the visual input provided by the user.

\begin{figure*}
    \centering
    \includegraphics[width=\linewidth]{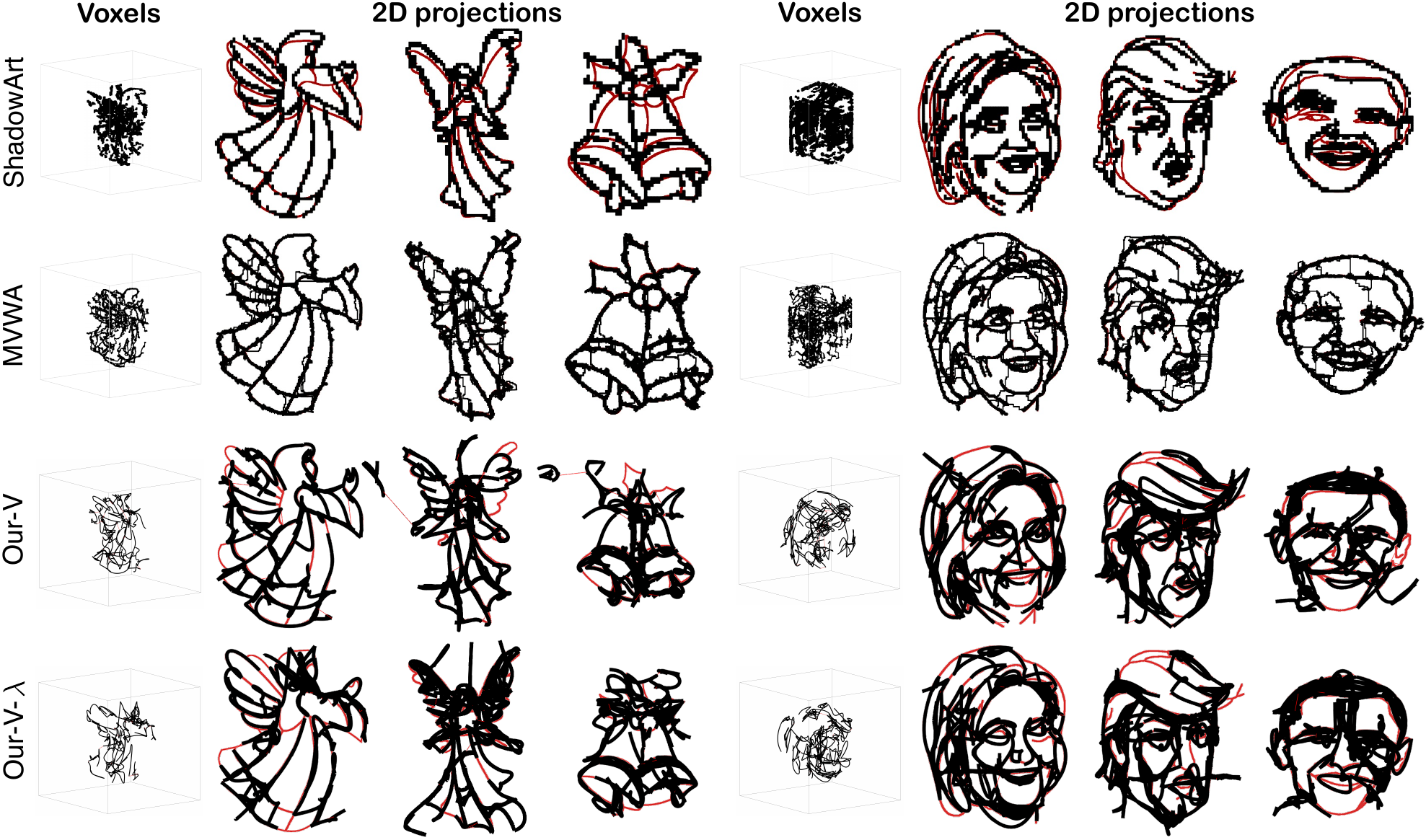}
    \caption{Comparison with existing multi-view wire art synthesis methods. The user-specified visual controls are highlighted with red lines.
    }
    \label{fig:qualitative_evaluation}
\end{figure*}

\begin{figure*}
    \centering
    \includegraphics[width=\linewidth]{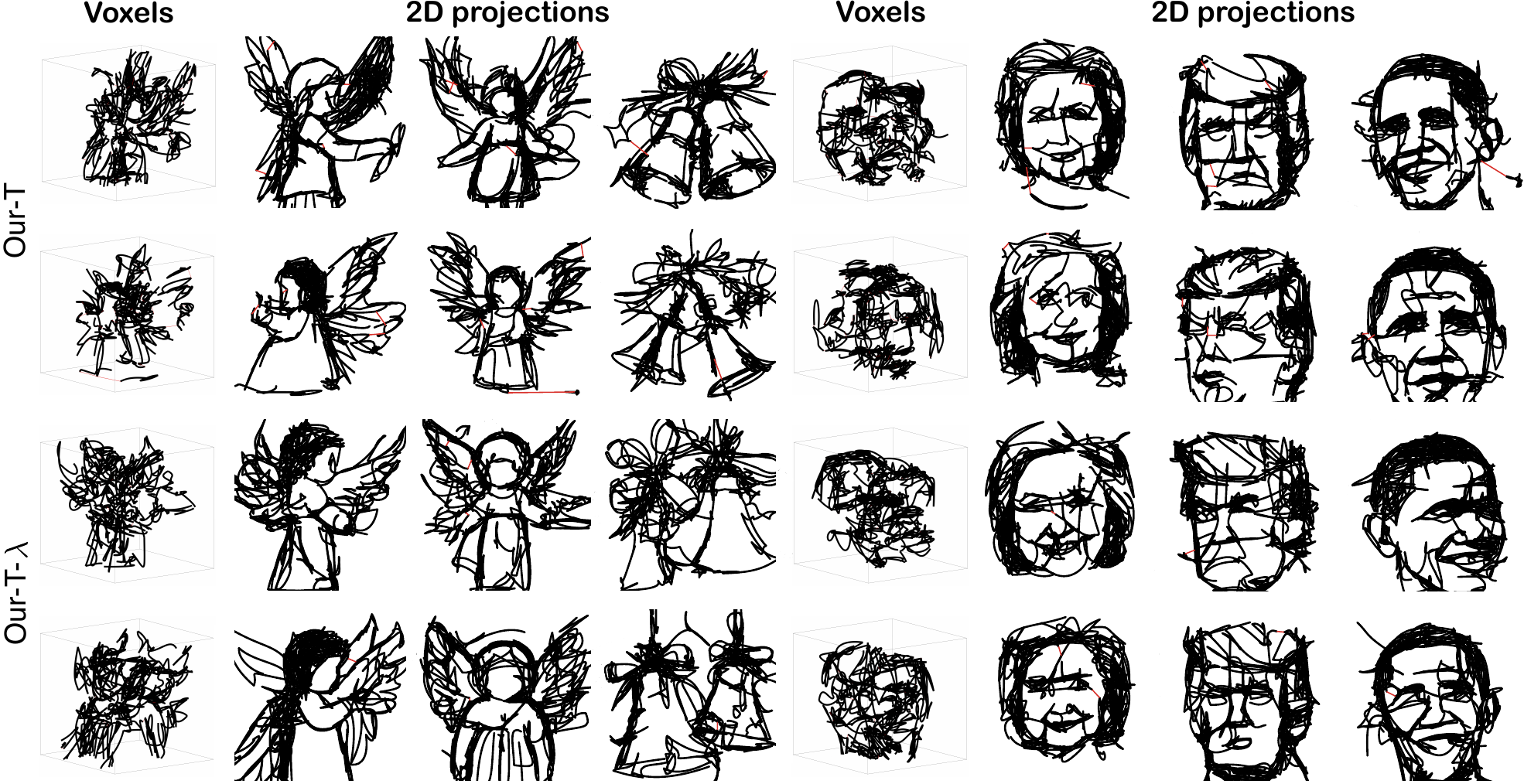}
    \caption{
    Additional instances of MVWA generated by our proposed \textit{\ours} with different random seeds. The text conditions for the MVWA on the left are defined as $\{c^X, c^Y, c^Z\}=\{$``a side view of an angel'', ``a front view of an angel'' ``Christmas bells''$\}$. Similarly, for the MVWA on the right are:  $\{c^X, c^Y, c^Z\}=\{$``Hillary'',``Trump'',``Obama''$\}$. The red lines indicate the additional connections that need to be added in order to form all the curves together as a whole. Compared to Our-T, Our-T-$\lambda$ ensures that the red lines are as short as possible while maintaining visual aesthetics.
    }
    \label{fig:ours_evaluation}
    \vspace{-0.8cm}
\end{figure*}

\subsection{Baseline comparison}
\label{sec:baseline}

We compared our approach against two state-of-the-art (SOTA) methods grounded in traditional graphics algorithms. \textbf{ShadowArt} \cite{shadowart} introduces a novel geometric optimisation method that automatically finds a consistent shadow hull by deforming the input images. \textbf{MVWA} \cite{hsiao2018multi} starts with reconstructing a discrete visual hull through intersecting generalised cones formed by back-projecting the given 2D image to 3D space and integrates the isolated components into a connected visual hull via a 3D path-finding method. \textbf{Our-V} and \textbf{Our-V-$\lambda$} are calibrated to align with the settings of these two methods, processing three user-specified line drawings alongside corresponding viewpoints as inputs. Here, $\lambda$ signifies the incorporation of MST regularisation.

The synthesised 3D wire arts, along with their corresponding 2D projections, are shown in Fig.~\ref{fig:qualitative_evaluation}. We can see that, the voxels yielded by ShadowArt \cite{shadowart} has a multitude of transformed components, attributable to the markedly inconsistent nature of the input line drawings, resulting in severely distorted 2D projections. Conversely, the MVWA \cite{hsiao2018multi} aspires to integrate the isolated components into a connected visual hull via a 3D path-finding method, inevitably incorporating numerous extraneous lines, thereby compromising the projected visuals' fidelity. In contrast, Our-V and Our-V-$\lambda$ are predicated upon the optimisation of a set of 3D B\'ezier curves, complemented by the utilisation of MST loss to emulate the impression of a singular line, a technique distinctly divergent from the conventional ``Voxel Hall Carving'' employed by preceding approaches. Our generated results are slightly inferior to the results of traditional methods in this setting as our approach cannot solve the input conflict problem very well. However, given only text as conditions, our method can generate projections with much higher quality (refer Fig.~\ref{fig:ours_evaluation}).
In addition, our results possess a more streamlined structural simplicity within the 3D space compared to traditional methods. 

\begin{table} [b]
\vspace{-0.2cm}
\centering
    \begin{adjustbox}{max width=\linewidth}
        \begin{tabular}{l c c c}
        \toprule 
         {\textbf{Methods}} & {\textbf{DINO-V2-Base}} & {\textbf{DINO-V2-Giant}} & {\textbf{CLIP Score}} \\ \midrule
        ShadowArt \cite{shadowart} & \textbf{79.62} & \textbf{83.82} & 34.37 \\
        MVWA \cite{hsiao2018multi} & 73.68 & 78.86 & 34.81\\
        Ours-V & 69.08 & 72.76 & 34.55\\
        Ours-V-$\lambda$ & 66.99 & 71.33 & 32.75\\
        \noalign{\smallskip}\cdashline{1-4}\noalign{\smallskip}
        Ours-T & - & - & \textbf{37.21}\\
        Ours-T-$\lambda$ & - & - & 36.52\\
        \bottomrule
        \end{tabular}
    \end{adjustbox}
    \vspace{-0.2cm}
    \caption{DINO similarity (\%) between the target sketches and the projection results and CLIP similarity (\%) between the captions and the projection results generated by different methods.}
    \label{tab:number}
\end{table}

\subsection{Main results} \label{subsec:main_results}
As demonstrated in Sec.~\ref{sec:baseline}, we have previously illustrated the capabilities of our method under \textbf{visual control}. In this section, we delineate the unique advantage of our approach over the baselines: our capacity to generate MVWA in response to textual input or a hybrid of text and visual inputs. This flexibility significantly diminishes the user's burden in resolving conflicts inherent in visual controls. 

\keypoint{Qualitative evaluation.} 
In addition to the examples shown in Fig.~\ref{fig:overview}, where we show the MVWA generated by our method under text and hybrid control, we further showcase massive cases in Fig.~\ref{fig:showtime} and Fig.~\ref{fig:ours_evaluation}. For each case, we present two variations created by \textbf{Our-T} and \textbf{Our-T-$\lambda$}. 
With MST regularisation, it becomes apparent that the resulting MVWA bears greater resemblance to coherent single-line 3D sculptures as opposed to an assemblage of numerous discrete visual hulls.

\keypoint{Quantitative evaluation.} 
The results are presented in Tab.~\ref{tab:number}. Our method does not achieve the highest DINO scores on several DINO variants. This is because our model fits the target sketches by continuously optimising the parameters of B\'ezier curves, compared to the search process of traditional methods, our optimisation method may suffer from underfitting and overfitting in different regions. Thus, given the layout constraints for various viewpoints, the traditional methods \cite{shadowart, hsiao2018multi} almost reach the best performance, and our method is still some distance away from them. However, given only the text conditions without layout constraints, our method enables high-quality fit to the target concept in numerous ways, omitting the step of the artist to elaborate the target projections. 

\begin{figure}[t]
    \centering
    \includegraphics[width=0.9\linewidth]{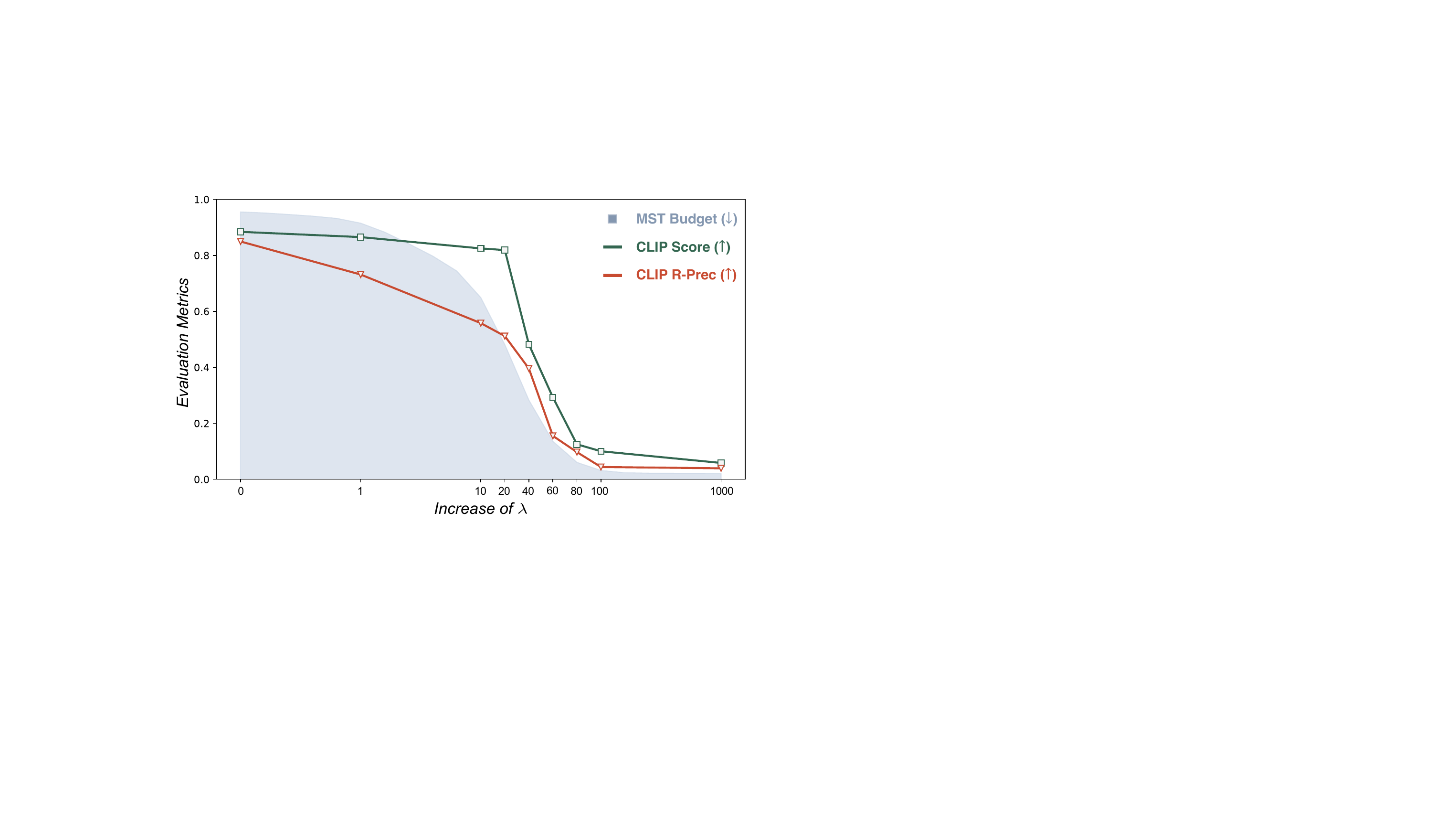}
    \vspace{-0.1cm}
    \caption{The impact of $\lambda$ on normalised evaluation metrics. Metrics are normalised to the [0, 1] interval for clarity. The original scale for MST budget spans [0, 8], for CLIP Score it is [36, 38], and for CLIP R-Prec it lies between [67, 85].}
    \label{fig:ablation}
    \vspace{-0.4cm}
\end{figure}

\keypoint{Ablation on $\lambda$.}
Fig.~\ref{fig:ablation} illustrates the influence of the  $\lambda$ coefficient within $\mathcal{L}_\text{MST}$ and CLIP metrics. The increase of $\lambda$ drives the generated MVWA towards an optimisation that favours a one-line wire art, leading to a substantial reduction in the wire connectivity budget. However, this also results in a notable decrease in both the CLIP-score and R-Precision, suggesting an increased deviation between the user input and the 2D projection of the synthesised MVWA. Taking into account both aesthetic appeal and manufacturability, we ultimately set the hyperparameter $\lambda$ to $50$.

\subsection{One Line vs. MST Regularisation}
\begin{figure}[t]
    \centering
    \includegraphics[width=\linewidth]{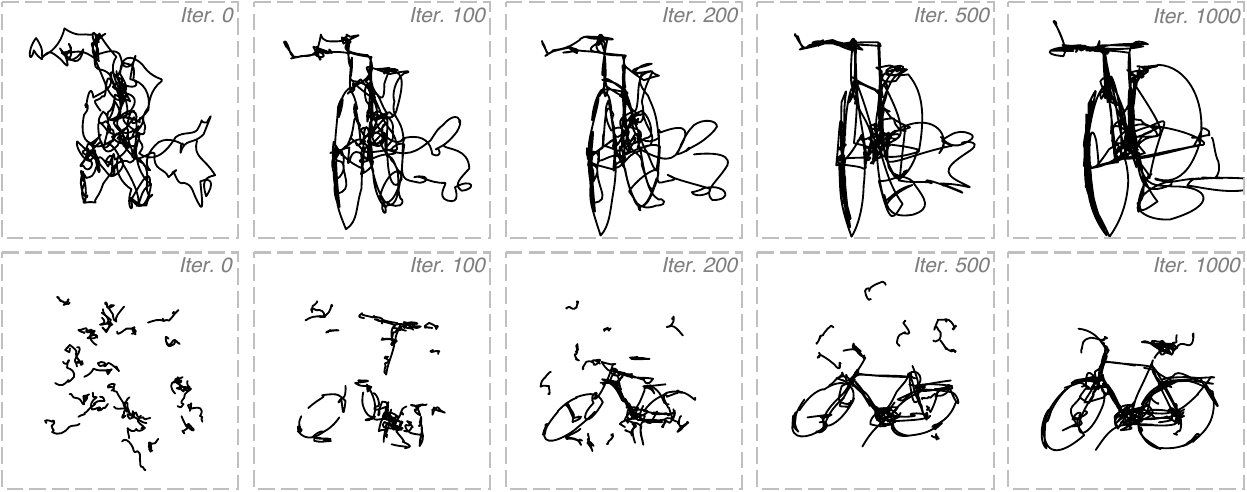}
    \caption{Generation of ``a simple drawing of a bicycle'' starts with a single curve containing 150 segments (top) and 30 curves containing 5 segments (bottom) in the same random seed.
    }
    \label{fig:oneline}
    \vspace{-0.4cm}
\end{figure}

In our endeavour to enhance the interconnectedness of generated 3D B\'ezier curves thereby more faithfully emulating a single, continuous curve, we have instituted a novel loss function, denoted as $\mathcal{L}_\text{MST}$. Intuitively, an alternative strategy could entail initialising the 3D wire art structure as a singular B\'ezier curve comprised of a substantially high segment count. Fig.~\ref{fig:oneline} (top) shows the generation result of ``a simple drawing of a bicycle'' when the input is a single B\'ezier curve with 150 segments. For a clearer presentation, only a 2D B\'ezier curve is used. It can be noted that a curve comprising numerous segments may not be able to update their positions effectively. This problem may lead to substantial segment overlap, thereby injecting redundancy and detracting from the succinctness of the ultimate generated form. Therefore, we utilise $\mathcal{L}_\text{MST}$ instead of the one-line setting to ensure that the wire art our model generates is aesthetically appealing and realistically producible.

\section{Discussion}

It is worth noting that \cite{vinker2022clipasso} emphasises the optimisation process is susceptible to the initialisation of B\'ezier curves. Therefore, the number, width and location of strokes and what text prompt to give each viewpoint may all be issues for artists to consider in future applications. In addition, compared to voxels and polylines, B\'ezier curves have fewer parameters, making them more advantageous for optimisation tasks. However, they may not fit targets well when the visual objectives contain a large number of non-smooth polylines. We anticipate the development of advanced line representation techniques in future research, which will enhance the fidelity of visual representations of MVWA.

\section{Conclusion}

This project is a pioneering venture into the fusion of AI and art, making strides by enabling AI to easily generate 3D multi-view wire art. It does so with just brief text prompts or spontaneous scribbles. Beyond its artistic impact, our work dives into scientific challenges related to abstraction and 3D representation in the generative AI community. The core of our approach involves refining 3D B\'ezier curves using diffusion models and a carefully designed rendering strategy. The ultimate goal is to make this distinct art form accessible to all, offering a platform for artists, designers, and enthusiasts to effortlessly bring their imaginative wire sculptures to life.

{\small
\bibliographystyle{ieee_fullname}
\bibliography{arxiv.bib}
}

\clearpage
\appendix

\onecolumn

\begin{center}
    {\Large\textbf{Supplementary Materials for\\Wired Perspectives: Multi-View Wire Art Embraces Generative AI}\par}
\end{center}

\section{Differentiable 3D MVWA rendering proof}
 
Any B\'ezier curve can be represented as the following format:
\begin{equation} \label{speq:bezier}
    B(t) = (1 - t)^3p_0 + 3(1 - t)^2t p_1 + 3(1 - t)t^2 p_2 + t^3 p_3.
\end{equation}

Given a plane $\pi$, its normal vector $N$ and a point $q$ in the plane, the projection of any point $P(t)$ of curve $B(t)$ on plane $\pi$ could be represented as $P^{\prime}(t)$:
\begin{equation} \label{speq:projection}
    P^{\prime}(t) = P(t) - (N \cdot (P(t) - q)) * N,
\end{equation}
\noindent where $\cdot$ denotes the dot product operation.

\vspace{1em}

Our objective is to prove that \textit{the projection of $B(t)$ on the plane $\pi$} ($B^{\prime}(t)$) is \textbf{equal} to \textit{the curve formed by the projected points of the control points of that curve} ($B^{\prime\prime}(t) = \{p_0^{\prime}, p_1^{\prime}, p_2^{\prime}, p_3^{\prime}\}$) on the same plane $\pi$, \ie, $B^{\prime}(t) = B^{\prime\prime}(t)$.

\vspace{1em}

Based on Eqs.~\ref{speq:bezier} and \ref{speq:projection}, we can obtain the projection curve $B^{\prime}(t)$ of $B(t)$ on the plane $\pi$:
\begin{equation} \label{speq:bezier'}
\begin{aligned}
    B^{\prime}(t) &= {\color{xred}[(1 - t)^3p_0 + 3(1 - t)^2tp_1 + 3(1 - t)t^2p_2 + t^3p_3]} \\
    &- {\color{xgreen}\{N \cdot [(1 - t)^3p_0 + 3(1 - t)^2tp_1 + 3(1 - t)t^2p_2 + t^3p_3 - q]\} * N}.
\end{aligned}
\end{equation}

We can also obtain the curve $B^{\prime\prime}(t)$ formed by the projected points of the control points on the plane $\pi$:
\begin{equation} \label{speq:bezier''}
\resizebox{\textwidth}{!}{
$\begin{aligned}
    B^{\prime\prime}(t) &= (1 - t)^3[p_0 - (N \cdot (p_0 - q)) * N] + 3(1 - t)^2t[p_1 - (N \cdot (p_1 - q)) * N] + 3(1 - t)t^2[p_2 - (N \cdot (p_2 - q)) * N] + t^3[p_3 - (N \cdot (p_3 - q)) * N] \\
    &= {\color{xred}[(1 - t)^3p_0 + 3(1 - t)^2tp_1 + 3(1 - t)t^2p_2 + t^3p_3]} \\
    &- {\color{xblue}[(1 - t)^3(N \cdot (p_0 - q)) * N + 3(1 - t)^2t(N \cdot (p_1 - q)) * N}{\color{xblue}\ + 3(1 - t)t^2(N \cdot (p_2 - q)) * N + t^3(N \cdot (p_3 - q)) * N]}.
\end{aligned}
$}
\end{equation}

Comparing Eq.~\ref{speq:bezier'} and \ref{speq:bezier''}, we find that the red parts are the same. Therefore, we only need to show that the \textcolor{xblue}{blue} part in Eq.~\ref{speq:bezier''} equals to the \textcolor{xgreen}{green} part in Eq.~\ref{speq:bezier'}. Since vector dot product satisfies the distributive over vector addition properties, \ie, $\Vec{a} \cdot (\Vec{b} + \Vec{c}) = \Vec{a} \cdot \Vec{b} + \Vec{a} \cdot \Vec{c}$, we can deform the blue part of Eq.~\ref{speq:bezier''} as follows:
\begin{equation} \label{speq:equal}
\begin{aligned}
    &\color{xblue}(1 - t)^3(N \cdot (p_0 - q)) * N + 3(1 - t)^2t(N \cdot (p_1 - q)) * N + 3(1 - t)t^2(N \cdot (p_2 - q)) * N + t^3(N \cdot (p_3 - q)) * N \\
    = &\{N \cdot [(1 - t)^3(p_0 - q) + 3(1 - t)^2t(p_1 - q) + 3(1 - t)t^2(p_2 - q) + t^3(p_3 - q)]\} * N \\
    = &\{N \cdot [(1 - t)^3p_0 + 3(1 - t)^2tp_1 + 3(1 - t)t^2p_2 + t^3p_3 - ((1 - t)^3 + 3(1 - t)^2t + 3(1 - t)t^2 + t^3) * q)]\} * N \\
    = &\color{xgreen}\{N \cdot [(1 - t)^3p_0 + 3(1 - t)^2tp_1 + 3(1 - t)t^2p_2 + t^3p_3 - q]\} * N
\end{aligned}
\end{equation}

Based on the above proof, we can transform the 3D B\'ezier curve rendering problem into a 2D rendering problem based on the projection of 3D control points. Given a 3D wire and a viewpoint, we can have a raster sketch with the utilisation of \cite{Li:2020:DVG} in our 3D to 2D renderer. And more importantly, the whole process is \textit{differentiable}.

\section{Bad case analysis}

When we use VectorFusion~\cite{jain2022vectorfusion} for 2D sketch generation, the results are limited by the diffusion prior. In general, given any text prompt, when the diffusion model generates images well, VectorFusion can also generate great sketches. However, this is not the case for {\ours}, as there are multi-view conflicts to consider. We analyse this problem with the following examples.

\begin{figure} [h!]
    \vspace{0.2cm}
    \centering
    \includegraphics[width=0.98\linewidth]{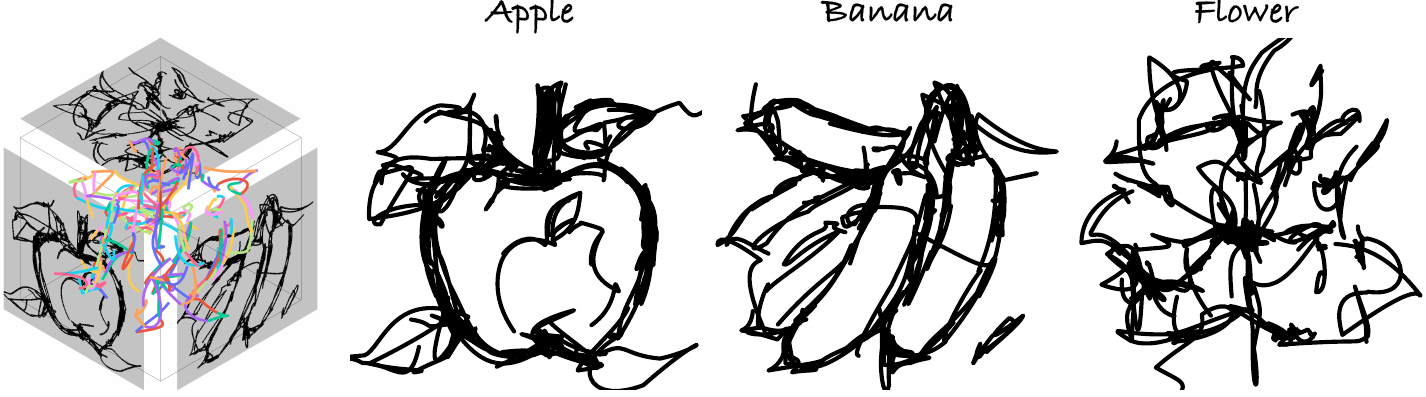}
    \vspace{0.3cm}
    \caption{The textual prompts for three viewpoints are ``apple'', ``banana'' and ``flower'', respectively. All prompts possess the following prompt prefix: ``a simple drawing of [text]''.} 
    \label{fig:apple}
    \vspace{-1em}
\end{figure}

Fig. \ref{fig:apple} shows the concepts of ``apple'', ``banana'' and ``flower''. However, we can notice that there is a lot of redundancy present in the sample of ``apple'', \eg, some leaves and a logo of ``apple'' in the centre. Compared to bananas and flowers, apples have a simpler structure. Therefore, if they are drawn within the same number of strokes, there are only two scenarios: the apples are drawn very intricately, or the bananas and flowers are drawn more abstractly. However, generative diffusion models like those in \cite{rombach2022high, shi2020improving}, trained solely on images, lack significant abstraction skills for simple stroke outlines. Consequently, \textit{\ours} is better at making complex sketches, \eg, ``portraits of three people''. This may go against human intuition. In addition, MVWA \cite{hsiao2018multi} mentioned that their method produces clearly visible artifacts due to the difficulty in resolving inconsistency from simple contours. Enhancing generative models' abstraction abilities could mitigate these challenges.

\begin{figure} [h!]
    \vspace{0.2cm}
    \centering
    \includegraphics[width=0.98\linewidth]{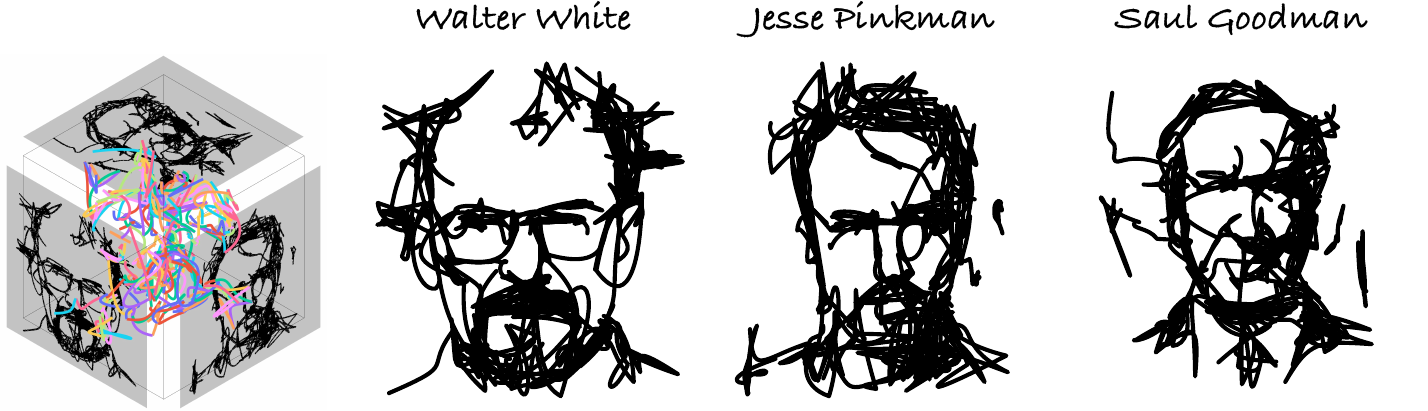}
    \vspace{0.3cm}
    \caption{The textual prompts for three viewpoints are ``Walter White'', ``Jesse Pinkman'' and ``Saul Goodman'', respectively. All prompts possess the following prompt prefix: ``a head of [text]''.} 
    \label{fig:walter}
    \vspace{-1em}
\end{figure}

Fig. \ref{fig:walter} shows the concepts of ``Walter White'', ``Jesse Pinkman'' and ``Saul Goodman''. We can notice that there is a lot of redundancy in the sample of ``Walter White''. It's common knowledge that Walter White is bald while Jesse Pinkman is not. During our generation process, their heads are positioned at the same height along the z-axis. Therefore, in order to draw Jesse's hair, \textit{\ours} has to add some extra strokes to the sideburns of Walter White. The potential for conflicting content of different viewpoints is an issue that users need to consider in future applications.

\section{Time-consuming analysis}

The time required to create a piece of Multi-View Wire Art varies significantly across different approaches: a professional artist typically requires several months. Compared to this, \textit{\ours} and the baseline method \cite{shadowart, hsiao2018multi} show a substantial improvement in time. ShadowArt \cite{shadowart} is based on the resize and voxel search of the target image, and it usually takes only 1 minute to complete. However, when there is a conflict between the three input target images (which is a very common situation), it produces a poor visual hull or even crashes. MVWA \cite{hsiao2018multi} takes about 2 to 10 hours, depending on the voxel resolution and the complexity of the input images, whereas \textit{\ours} significantly reduces this to approximately 30 minutes. Our method utilises a text-to-image generation model to raise the upper limit of creativity with acceptable time consumption compared to rule-based methods. While SDS may not be recognised for its efficiency in AIGC, it emerges as the most efficient method known to us for the task of Multi-View Wire Art creation. In addition, with the widespread use of SDS for 3D generation tasks, the acceleration of SDS has been investigated \cite{zhou2023dreampropeller}. We anticipate that the performance of our proposed method will be enhanced with advancements in generative AI.

\section{Physical display}

\begin{figure} [h!]
    \vspace{0.2cm}
    \centering
    \includegraphics[width=\linewidth]{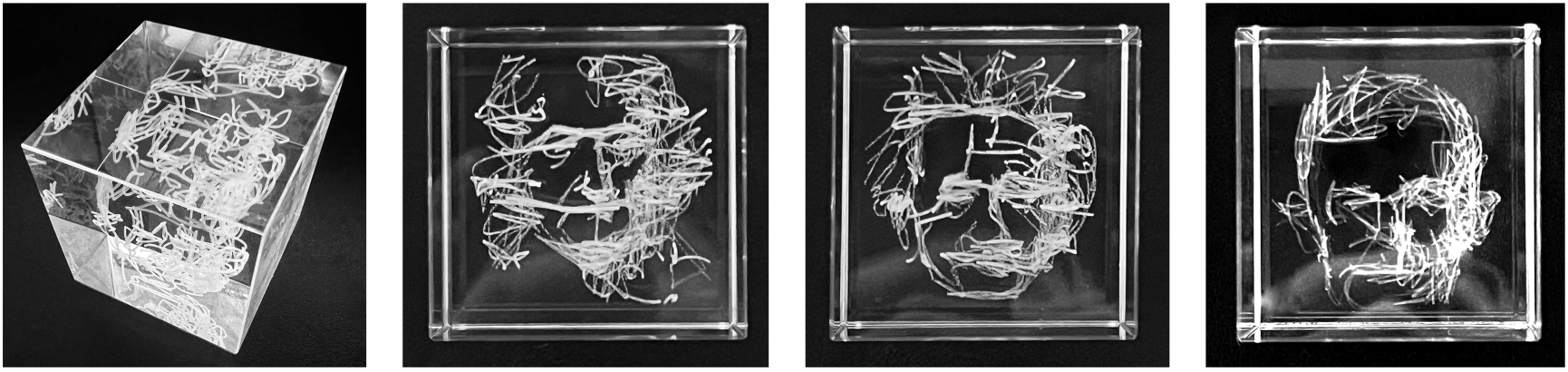}
    \caption{The textual prompts for three viewpoints are ``Isaac Newton'', ``Albert Einstein'' and ``Alan Turing'', respectively. All prompts possess the following prompt prefix: ``a head of [text]''.} 
    \label{fig:crystal}
\end{figure}

In our main paper, we highlight that utilising $\mathcal{L}_\text{MST}$ compromises aesthetics. To address this, we employ laser crystal in creating our multi-view wire art, as demonstrated in Fig.~\ref{fig:crystal}. Please watch the \href{https://www.youtube.com/watch?v=8yCUGgnO4vY}{video} to experience the fun of changing perspectives.

\end{document}